\documentclass[10pt,twocolumn,letterpaper]{article}

\usepackage{cvpr}              

\usepackage{graphicx}
\usepackage{amsmath}
\usepackage{amssymb}
\usepackage{booktabs}
\usepackage{multirow}
\usepackage{comment}
\usepackage{color}
\usepackage[normalem]{ulem}
\usepackage{paralist, tabularx}
\definecolor{grey}{RGB}{128,138,135}

\usepackage[pagebackref,breaklinks,colorlinks]{hyperref}
\newcommand\blfootnote[1]{%
\begingroup
\renewcommand\thefootnote{}\footnote{#1}%
\addtocounter{footnote}{-1}%
\endgroup
}
\usepackage[accsupp]{axessibility}

\usepackage[capitalize]{cleveref}
\crefname{section}{Sec.}{Secs.}
\Crefname{section}{Section}{Sections}
\Crefname{table}{Table}{Tables}
\crefname{table}{Tab.}{Tabs.}

\def\cvprPaperID{8811} 
\def\confName{CVPR}
\def\confYear{2022}

\begin{document}

\title{Learning to Prompt for Open-Vocabulary Object Detection with Vision-Language Model}

\author{Yu Du$^{1}$ \quad Fangyun Wei$^{2\dagger}$ \quad Zihe Zhang$^{1}$ \quad Miaojing Shi$^{3\dagger}$ \quad Yue Gao$^2$ \quad Guoqi Li$^1$ \vspace{4pt}\\
	$^1$Tsinghua University \quad
    $^2$Microsoft Research Asia  \quad 
    $^3$King’s College London \\
     {\tt\small \{duyu20, zh-zhang17\}@mails.tsinghua.edu.cn} \quad
    {\tt\small liguoqi@mail.tsinghua.edu.cn} \\
	{\tt\small \{fawe, yuegao\}@microsoft.com} \quad {\tt\small miaojing.shi@kcl.ac.uk} \\
}

\maketitle

\begin{abstract}
Recently, vision-language pre-training shows great potential in open-vocabulary object detection, where detectors trained on base classes are devised for detecting new classes. The class text embedding is firstly generated by feeding prompts to the text encoder of a pre-trained vision-language model. It is then used as the region classifier to supervise the training of a detector. The key element that leads to the success of this model is the proper prompt, which requires careful words tuning and ingenious design. To avoid laborious prompt engineering, there are some prompt representation learning methods being proposed for the image classification task, which however can only be sub-optimal solutions when applied to the detection task. In this paper, we introduce a novel method, detection prompt (DetPro), to learn continuous prompt representations for open-vocabulary object detection based on the pre-trained vision-language model. Different from the previous classification-oriented methods, DetPro has two highlights: 1) a background interpretation scheme to include the proposals in image background into the prompt training; 2) a context grading scheme to separate proposals in image foreground for tailored prompt training. We assemble DetPro with ViLD, a recent state-of-the-art open-world object detector, and conduct experiments on the LVIS as well as transfer learning on the Pascal VOC, COCO, Objects365 datasets. Experimental results show that our DetPro outperforms the baseline ViLD~\cite{gu2021zero} in all settings, e.g., +3.4 AP$^{\text{box}}$ and +3.0 AP$^{\text{mask}}$ improvements on the novel classes of LVIS. Code and models are available at \url{https://github.com/dyabel/detpro}. 
\end{abstract}
\blfootnote{ $^{\dagger}$Corresponding author.}
\vspace{-2mm}
\section{Introduction}
\label{sec:intro}
Object detection aims at locating bounding boxes of objects in an image as well as assigning labels to them. In last few years, object detection~\cite{ren2015faster,redmon2016you} achieves great success in solving the closed-set problem, \ie, detectors can detect classes present in the training set. To increase the detection vocabulary, the common practice is by collecting more data with desired classes. Besides the expensive labeling cost in this process, it often leads to a long-tailed distribution~\cite{gupta2019lvis,li2020overcoming} of object classes: detectors need to be carefully designed to avoid overfitting on frequently-occurred categories in the dataset. In contrast, an alternative way for increasing the detection vocabulary is open-vocabulary object detection (OVOD), where detectors are trained on base classes and equipped with ability to detect new classes.

Recently, ViLD~\cite{gu2021zero} introduces a framework for open-vocabulary object detection, which distills the knowledge from a pre-trained vision-language model into a detector. It is inspired by the recent progress of vision-language pre-training, e.g, CLIP~\cite{radford2021learning} and ALIGN~\cite{jia2021scaling}, where two separate encoders, namely image encoder and text encoder, are used to maximize the alignment between images and corresponding texts. In ViLD's implementation, they feed text descriptions of base classes,
known as prompt, into the text encoder of CLIP to generate the class text embedding. The embedding is then utilized to classify object proposals and supervise the detector training. To perform open-set object detection, the base class text embedding is replaced with the embedding of both base and novel classes.
The prompt design, also known as prompt engineering, is crucial in this process as we observe a slight word change in it would end up with {clear positive or negative impact on the detection performance}. Designing proper prompts requires domain expertise and carefully word tuning from human, as of~\cite{gu2021zero}. To avoid such high-end and rather laborious demand from human, the alternative way is to automatically learn prompt’s context using continuous representations, we name it as prompt representation learning in our work.

In this paper, we propose a novel method named detection prompt (DetPro) to learn prompt representations, in the setting of open-vocabulary object detection with pre-trained vision-language model (OVOD-VLM). There are some recent works focusing on prompt representation learning such as CoOp~\cite{zhou2021learning}, who targets for improving image classification accuracy based on the pre-trained vision-language models. Directly applying CoOP into the OVOD-VLM is not realistic: 
image classification only needs to recognize the correct labels of input images while object detection requires detectors to distinguish foregrounds from backgrounds, and classify region proposals in foregrounds into different object classes. We thus introduce a new Detection Prompt (DetPro) to automatically learn prompt representations in OVOD-LVM based on positive and negative proposals \wrt ground truth in images.

Prompt learning in object detection faces two critical issues: 1) Negative proposals, despite being very important to object detection, do not correspond to specific object classes, therefore can not be easily included into the prompt learning process. 2) Unlike objects in image classification being centered and big in images, objects in positive proposals are often associated with different levels of contexts, learning one prompt context for these proposals can not be sufficient. To tackle them, we introduce,
\begin{compactitem}
\item  a background interpretation scheme for negative proposal inclusion, which optimizes the embedding of negative proposals to be away from all other class embedding; 
\item  a context grading scheme with tailored positive proposals, which tailors the prompt representation learning with different positive proposal sets corresponding to different context levels.   
\end{compactitem}
We assemble DetPro  with ViLD~\cite{gu2021zero}, and conduct a series of experiments on LVIS and transfer the LVIS-trained model to other datasets including Pascal VOC, COCO and Objects365. In all settings, our DetPro outperforms the ViLD, e.g., +3.4 AP$^{\text{box}}$ and +3.0 AP$^{\text{mask}}$ improvements on the novel classes of LVIS.

\section{Related Work}
\label{sec:related}

\noindent \textbf{Prompt Learning.} Recently, the development of large vision-language model (VLM), \eg, CLIP~\cite{radford2021learning} and ALIGN~\cite{jia2021scaling}, emerges and finds its applications in few-shot or zero-shot learning tasks~\cite{ule2003clip,fang2021clip2video}.
The VLMs are trained on huge amount of image-text pairs collected from web and contrastive learning~\cite{khosla2020supervised} is adopted to align the image and text embedding.
The pretrained VLMs can be transferred to its downstream tasks with either finetuning~\cite{lu2019vilbert,su2019vl} or prompt engineering~\cite{zhou2021learning}.
A task-specific prompt can boost the performance significantly~\cite{radford2021learning} but requires laborious prompt engineering. Inspired by prompt learning in language tasks, CoOp~\cite{zhou2021learning} proposes the context optimization to automate prompt engineering for few-shot classification. It models the context of prompts as continuous representations that are end-to-end learned from a small set of data.
This paper extends CoOP to OVOD by designing special strategies to handle foreground and background proposals within images. While CoOP learns the prompt with samples of all categories our DetPro is trained on only base classes and expected to generalize to novel classes.

\noindent \textbf{Open-Vocabulary Object Detection.}
  Despite the remarkable success of DNNs~\cite{krizhevsky2012imagenet,simonyan2014very,he2016deep,cao2020global} in the computer vision field, they often require a large amount of annotated data in order to get satisfying results of object detection~\cite{ren2015faster,lin2017focal,wei2020point,chi2020relationnet++}.
  To alleviate the reliability of DNNs on big data and elaborate annotations, different paradigms such as semi-supervised learning~\cite{xu2021end}, few-shot learning~\cite{snell2017prototypical,sung2018learning,yang2020restoring}, zero-shot learning~\cite{romera2015embarrassingly,wang2019survey,xian2018zero}, self-supervised learning~\cite{wei2021aligning}, open-set learning~\cite{vyas2018out,geng2020recent,scheirer2014probability} and advanced training strategies~\cite{xu2021bootstrap,ming2019group} are introduced.   

Particularly, for the zero-shot detection task, it aims to generalize from seen classes (with bounding box annotations) to unseen classes. Despite they have made some progress, their overall performance is still far behind the fully-supervised methods~\cite{zhu2019zero,bansal2018zero}, therefore research on it is not flourishing yet.

Recently, open-vocabulary object detection emerges as a more general and practical paradigm onto the stage than the zero-shot detection: an unbounded vocabulary of concepts is firstly acquired by training on image-text pairs, then the detector is required to detect novel classes with the availability of bounding box annotations of a number of base classes. The representative solutions include OVR-CNN ~\cite{zareian2021open} and ViLD~\cite{gu2021zero}. OVR-CNN~\cite{zareian2021open} pretrains the backbone using a corpus of image-caption pairs and finetunes the detector with annotations of only a few object categories while ViLD~\cite{gu2021zero} directly distills knowledge from a pretrained open-vocabulary classification model into a two-stage detector.

\begin{figure*}[t]
    \centering
    \includegraphics[width=0.98\textwidth]{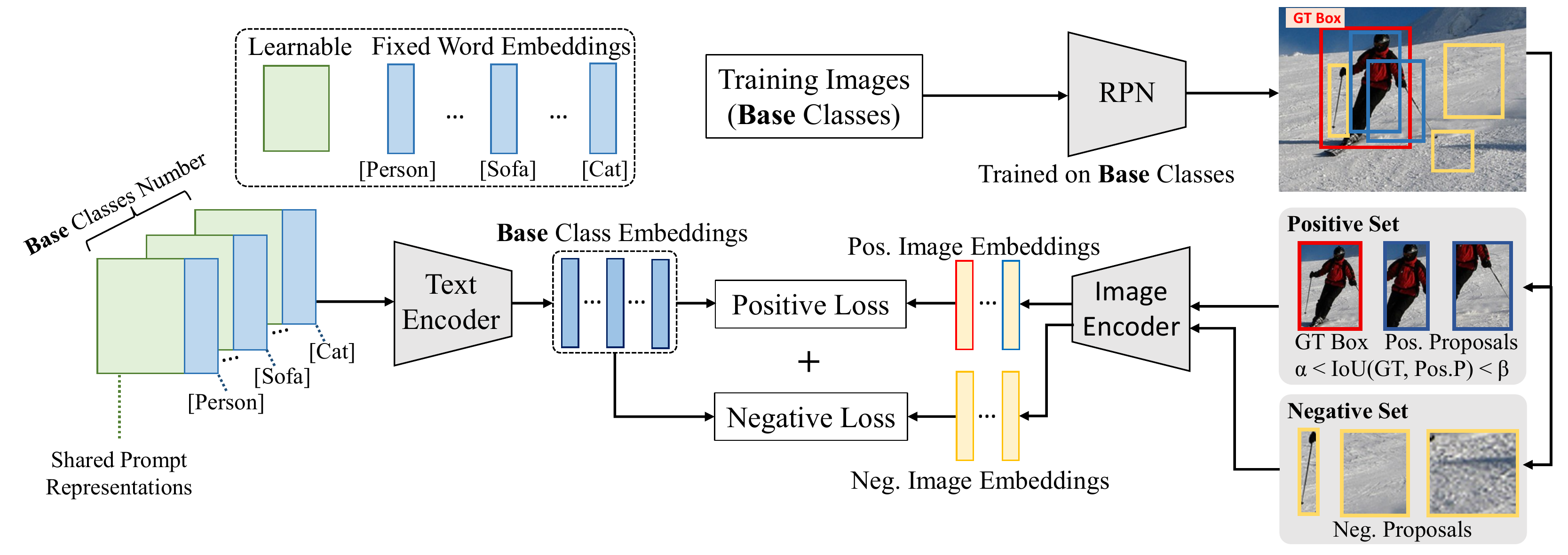}
    \vspace{-2mm}
    \caption{Overview of DetPro. Positive loss is defined between visual embedding of positive proposals in images and their corresponding class embedding; while negative loss is defined between visual embedding of negative proposals and all class embedding. Different tailored positive proposal sets ($\alpha < \text{IoU(GT, Pos P)} < \beta$) are used to learn different prompt representations and are ensemabled in the end. }
    \label{fig:framework}
    \vspace{-3mm}
\end{figure*}
We place our work in the OVOD setting and build our solution upon the ViLD~\cite{gu2021zero}. ViLD uses hand-crafted prompts for generating class embedding, while we design fine-grained automatic prompt learning and special background interpretation to find the desired prompts.

\section{Problem Setting}
\label{sec:method}

The goal of DetPro is to learn continuous prompt representations for OVOD-VLM. Figure~\ref{fig:framework} shows an overview of our DetPro, which includes two key elements: background interpretation for negative proposal inclusion, foreground context grading with tailored positive proposals. They are dedicated to the positive and negative losses in the figure. Afterwards, we devise DetPro upon the recent OVOD pipeline ViLD~\cite{gu2021zero} in Figure~\ref{fig:vild} where DetPro serves as a replacement for the proposal classifier in ViLD to realize automatic prompt engineering for it.

\noindent\textbf{Data Split.} We divide categories in a detection dataset into two disjoint sets for base classes $\mathcal{C}_B$ and novel classes $\mathcal {C}_N$. We use $|\mathcal{C}_B|$ and $|\mathcal{C}_N|$ to denote the number of base and novel classes, respectively. Correspondingly, we have $\mathcal{X}_{T}$ and $\mathcal{X}_{I}$ for the training and inference dataset, respectively. $\mathcal{X}_{T}$ contains only base classes $\mathcal{C}_B$ with annotations for training, while $\mathcal{X}_{I}$ contains both $\mathcal{C}_B$ and $\mathcal{C}_N$ for the trained model to recognize objects from both $\mathcal {C}_B$ and $\mathcal{C}_N$.

\noindent\textbf{Pre-trained Vison-language Model.} We use CLIP~\cite{radford2021learning} as our vision-language model, which consists of a text encoder $\mathcal{T}(\cdot)$ and an image encoder $\mathcal{I}(\cdot)$. $\mathcal{T}(\cdot)$ takes the input of prompt representation of a class and outputs the corresponding text embedding, which is also named as class embedding in our work; $\mathcal{I}(\cdot)$ takes the input of an image of size $224 \times 224$ and outputs the corresponding image embedding.

\noindent\textbf{Detection Framework.} We adopt Faster-RCNN with ResNet-50 and FPN as our detector.

\section{Method}
We first review prompt representation learning in image classification, then we present our DetPro in object detection; finally, assemble it onto ViLD for OVOD.
\label{sec:detpro}
\vspace{-1mm}
\subsection{\bf Preliminaries: Prompt}
Original CLIP~\cite{radford2021learning} feeds human-defined prompt, \eg `a photo of [CLASS]', into its text encoder $\mathcal{T}$ to generate the class embedding for image classification. In a specific case, [CLASS] is replaced by the class name such as `person' and `cat'. 
Identifying the proper prompt is a non-trivial task, which often costs a significant amount of time for words tuning. To bypass it, CoOp~\cite{zhou2021learning} proposes to automatically learn prompt representations. The learnable prompt representation $\boldsymbol V_c$ for given class $c \in C_{B}$ is defined as follows:
\vspace{-3mm}
\begin{equation}
  \boldsymbol V_c =[\boldsymbol v_1,\boldsymbol v_2\dots\boldsymbol v_L,\boldsymbol w_{{c}}],
\label{eq:ctx1}
\end{equation}
where $\boldsymbol{v}_i$ denotes the $i$-th leanable context vector, $\boldsymbol w_{c}$ the fixed class token of base class $c$ and $L$ the context length. $[\boldsymbol v_1,\boldsymbol v_2\dots\boldsymbol v_L]$ can be analogue to context of the human-defined prompt, \eg `a photo of', while $\boldsymbol w_{{c}}$ analog to the class name [CLASS].  
$\{\boldsymbol v_i\}_{i=1}^L$ are randomly initialized to have the same dimension to the word embedding $\boldsymbol w_{{c}}$ (512 in this work). The learned prompt context $[\boldsymbol v_1,\boldsymbol v_2\dots\boldsymbol v_L]$ is shared across classes, such that when a new class comes, its prompt representation can be easily obtained by (\ref{eq:ctx1}). 
The class embedding $\boldsymbol t_c$ of class $c$ is generated by feeding $\boldsymbol V_c$ into the CLIP text encoder $\mathcal{T}(\cdot)$:
\begin{equation}
\label{eq:clip_text}
\boldsymbol t_c=\mathcal{T}(\boldsymbol V_c).
\end{equation}

In image classification task, given an image $x$, we can first feed it into the CLIP image encoder $\mathcal{I}(\cdot)$ to extract its image embedding $\boldsymbol f$. Assuming this image belongs to class $c$, the probability of $\boldsymbol f$  being classified as class $c$ is computed as:
\begin{equation}
\label{eq:postive_prob}
    p_c = \frac{\exp(\cos(\boldsymbol f,\boldsymbol t_c)/\tau)}{\sum_{i\in \mathcal {C}_B}\exp(\cos(\boldsymbol f,\boldsymbol t_{i})/\tau)},
\end{equation}
where $\tau$ is a temperature parameter, $\cos(\cdot,\cdot)$ denotes the cosine similarity. The cross entropy loss is applied to optimize $[\boldsymbol v_1,\boldsymbol v_2\dots\boldsymbol v_L]$ while both $\mathcal{I}(\cdot)$ and $\mathcal{T}(\cdot)$ are fixed:
\begin{equation}
\label{eq:pos_loss}
    \mathcal{L}_{p} = -\log p_c.
\end{equation}

\vspace{-2mm}
\subsection{Detection Prompt}
\label{sec:detection_prompt}
\noindent{\bf Na\"ive Solution.} Object detection differs from image classification as for each training image we have class labels provided on ground truth bounding boxes of objects, and for each test image we need to localize bounding boxes of objects and predict class labels for them. To adapt the prompt representation learning strategy CoOP~\cite{zhou2021learning} into the detection task, the straightforward way is to simulate the classification scenario that it works: given an image $x$, we instead feed its cropped ground truth bounding boxes into the CLIP image encoder $\mathcal{I}(\cdot)$ to obtain the box embedding $\boldsymbol f$, respectively. Each ground truth box belongs to only one object class $c$; we denote by $\mathcal G$ all ground truth bounding boxes over images. We can then follow the same equations (\ref{eq:postive_prob},\ref{eq:pos_loss}) to learn a region-level classifier on $\mathcal G$. This classifier can be further assembled with an established object detection pipeline (\eg Faster R-CNN), specified in Section~\ref{sec:apply}. 

This na\"ive adaption can work to certain extent, but is only a sub-optimal solution: the rich information in images apart from the ground truth bounding boxes has been dropped, including foreground and background proposals, this however is essential to learn a robust region-level (proposal) classifier for detection.  

\noindent{\bf Fine-grained Solution.} In order to make use of image proposals, we first train a RPN on base classes $\mathcal{C}_B$ to extract them from $\mathcal X_T$. Foreground proposals $\mathcal F$ are those whose IoUs \wrt one ground truth in $\mathcal G$ are larger than a thresh, \ie 0.5,  while background proposals $\mathcal B$ are negative proposals whose IoUs \wrt all ground truth in $\mathcal G$ are smaller than the thresh. We call the union of $\mathcal F$ and $\mathcal G$ the positive proposal set $\mathcal P$, \ie $\mathcal P = \mathcal F ~\cup~ \mathcal G$, and $\mathcal B$ the negative proposal set $\mathcal N$, \ie $\mathcal N = \mathcal B$. 
For a proposal in $\mathcal P$, unless it's the ground truth whose target object inside is tightly bounded, it normally includes a big partial of the object with considerable surrounding context. The positive proposals thus vary a lot in contexts depending on their IoUs \wrt the ground truth. This shall result into different visual embedding when feeding them into $\mathcal{I}(\cdot)$. Consequently, different prompt representations should also be learned dedicated to different prompt contexts. 
To address this issue, we introduce a {context grading scheme with tailored positive proposals} (specified later). On the other hand, for a proposal in $\mathcal N$, it contains mostly background with the possibility of a small partial of target objects. The background does not have a specific class name, thus its prompt representation can not be directly obtained (no $\boldsymbol w_{{c}}$ in Eq.\ref{eq:ctx1}), nor does its class embedding. Negative proposals serve as a very important role in object detection. In order to utilize them in our detection prompt, we introduce a {background interpretation scheme for negative proposal inclusion}. Below we detail it.          

\noindent{\emph{Background interpretation for negative proposal inclusion.}} Background might contain some object classes inside, but they can not be normally recognized as consequences of being either too small, too incomplete or too vague. In other words, given an negative proposal $n$, its image embedding  $\boldsymbol f_n$ by $\mathcal{I}(\cdot)$ should be dissimilar to any text embedding $\boldsymbol t_c$ of other classes by $\mathcal{T}(\cdot)$. 

The probability $p_{nc}$ of $\boldsymbol f_n$ being classified as class $c$ is computed via Eq.\ref{eq:postive_prob}. We want any $p_{nc}$ to be small;  in practice, since $|\mathcal C_B|$ is big, we could simply optimize any $p_{nc}$ to $\frac{1}{|\mathcal C_B|}$. This forces the negative proposal to be equally unlike any object classes. The loss function is thus formulated as, 
\begin{equation}
    \mathcal{L}_{\text{n}}
    =-\sum_{c=1}^{|\mathcal C_B|} w\log p_{nc},~~~~w = \frac{1}{|\mathcal C_B|}.
\label{eq:softBG}
\end{equation}

An alternative way for background interpretation is to learn a stand alone background prompt representation $\boldsymbol V_{\text{bg}}$ which is similar to $\boldsymbol V_{c}$ for class $c$ but without the class token:
\begin{equation}
    \boldsymbol V_{\text{bg}} =[\boldsymbol v^{\text{bg}}_1\boldsymbol, \boldsymbol v^{\text{bg}}_2,\dots,\boldsymbol v^{\text{bg}}_{L}].
\label{eq:ctx}
\end{equation}
Similarly, we use Eq.\ref{eq:clip_text} to generate background embedding $\boldsymbol{t}_{\text{bg}}$ and feed the negative proposal $n$ into $\mathcal{I}(\cdot)$ to generate $\boldsymbol f_n$. The probability $p_{nbg}$ is computed as:
\begin{equation}
    p_{\text{nbg}} = \frac{\exp{(\cos(\boldsymbol{f_n},\boldsymbol{t}_{\text{bg}})/\tau)}}{\sum_{c=1}^{\mathcal{C}_B}\exp{(\cos(\boldsymbol{f_n},\boldsymbol t_c)/\tau)} +\exp{(\cos(\boldsymbol{f_n},\boldsymbol{t}_{bg})/\tau)} }
    \label{eq:neg_prob}.
\end{equation}
The negative loss is defined as:
\begin{equation}
    \mathcal{L}_{\text{n}} = -\log p_{\text{nbg}}.
    \label{eq:negloss}
\end{equation}
This alternative way is inferior to the first way. The background content may vary a lot, the second way learns an explicit background embedding to let all negative proposals be close to it, which can not be sufficient. In contrast, in the first way it is implicitly interpreted to let each negative proposal being away from all other class embedding, which can be more robust.

\begin{figure*}
    \centering
    \includegraphics[width=0.98\textwidth]{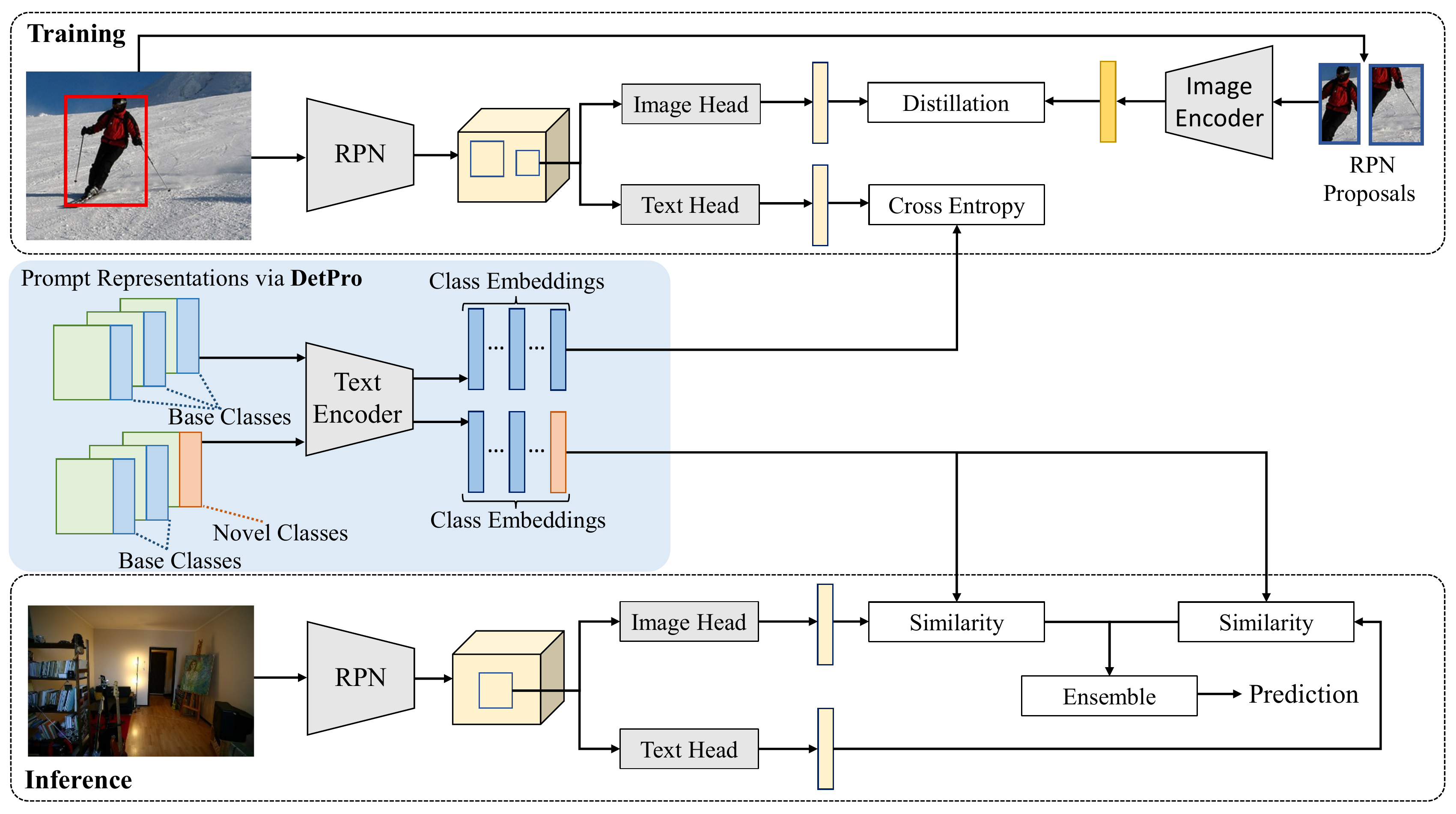}
    \vspace{-2mm}
    \caption{Assembling DetPro with ViLD. DetPro is highlighted with azure background. We omit the class-agnostic bounding box regression branch and mask prediction branch in both training and testing pipelines.}
    \label{fig:vild}
    \vspace{-4mm}
\end{figure*}

\noindent{\emph{Context grading with tailored positive proposals.}} A positive proposal may incorporate different contexts \wrt the target object. This difference can be analogue in the prompt context: given a ground truth bounding box of an object class, we can say `a photo of [CLASS]'; while given a foreground proposal of a partial object, we could instead say `a photo of partial [CLASS]'. The learned prompt context representations for `a photo of' and `a photo of partial' will be different, which ends up with different class embedding for the two types of prompts. They should be optimized with positive proposals corresponding to different levels of contexts, respectively. We introduce a foreground context grading scheme with tailored positive proposals for this purpose. 

Specifically, we divide the positive proposals of the IoU range $[a,b]$ into $K$ disjoint groups with an IoU interval of $t$, such that $K=(a-b)/t$. The foreground context will be graded in different groups, such that positive proposals within each group have similar context level \wrt their respective ground truth. We therefore learn prompt representations in the $K$ groups independently. Within the $k$-th group, we follow the same equations (\ref{eq:postive_prob},\ref{eq:pos_loss}) to extract visual embedding $\boldsymbol f_p$, compute probability $p_{pc}$, and optimize positive loss  $\mathcal L_p$ for any positive proposal $p$ of class $c$ inside. The same negative proposal set $\mathcal N$ is included into each group such that the final loss function within each group is, 
\begin{equation}
    \mathcal{L} = \frac{1}{|\mathcal N|} \sum_{n  \in \mathcal N} \mathcal L_n + \frac{1}{|\mathcal P^k|} \sum_{p  \in \mathcal P^k} \mathcal L_p.    \label{eq:finalloss}
\end{equation}
The prompt representation $\boldsymbol V_c^k$ is learned in each group for class $c$. In the end, the learned representations are ensembled over $K$ groups by average, such that $\boldsymbol V_c = \frac{1}{K}\sum_{k=1}^K \boldsymbol V_c^k$.

\subsection{Assembling DetPro onto ViLD}
\label{sec:apply}
ViLD~\cite{gu2021zero} is a recent framework for OVOD. It distills the knowledge from CLIP~\cite{radford2021learning} into a two stage detector, \ie, Faster R-CNN~\cite{ren2015faster}. Figure~\ref{fig:vild} shows assembling our DetPro with ViLD.

\noindent\textbf{Training ViLD with DetPro.} A learned DetPro generates prompt representations based on Eq.~\ref{eq:ctx1} for base classes, which we can feed into $\mathcal{T}(\cdot)$ to generate base class embedding. The embedding is used as proposal classifier for the detector. Following ViLD, we employ two R-CNN heads (sub-branches), namely image head and text head. The image head distills knowledge from CLIP image encoder, while the text head replaces the original R-CNN classifier by our base class embedding (fixed) plus a learnable background embedding (see Figure~\ref{fig:vild}).

\begin{table*}[t]
  \small
  \centering
  \begin{tabular}{l|c|cccc|cccc}
  \toprule
 \multirow{2}{*}{Method}&\multirow{2}{*}{Epoch}&\multicolumn{4}{c|}{Detection}&\multicolumn{4}{c}{Instance segmentation}\\
  &&AP$_{r}$&\textcolor{grey}{AP$_{c}$}&\textcolor{grey}{AP$_{f}$}&\textcolor{grey}{AP}&AP$_{r}$&\textcolor{grey}{AP$_{c}$}&\textcolor{grey}{AP$_{f}$}&\textcolor{grey}{AP}\\
  \midrule
  Supervised (base)& 20 &0.0&\textcolor{grey}{26.1}&\textcolor{grey}{34.0}&\textcolor{grey}{24.7}&0.0&\textcolor{grey}{24.7}&\textcolor{grey}{29.8}&\textcolor{grey}{22.4}\\
  Supervised (base+novel)& 20 &15.5&\textcolor{grey}{25.5}&\textcolor{grey}{33.6}&\textcolor{grey}{27.0}&16.4&\textcolor{grey}{24.6}&\textcolor{grey}{30.6}&\textcolor{grey}{25.5}\\
  \midrule
  ViLD (base)\cite{gu2021zero} &460 &16.7&\textcolor{grey}{26.5}&\textcolor{grey}{34.2}&\textcolor{grey}{27.8}&16.6&\textcolor{grey}{24.6}&\textcolor{grey}{30.3}&\textcolor{grey}{25.5}\\
  ViLD* (base)~\cite{gu2021zero}  & 20 &17.4&\textcolor{grey}{27.5}&\textcolor{grey}{31.9}&\textcolor{grey}{27.5}&16.8&\textcolor{grey}{25.6}&\textcolor{grey}{28.5}&\textcolor{grey}{25.2}\\
  DetPro (base) & 20 & \textbf{20.8} &\textcolor{grey}{27.8}&\textcolor{grey}{32.4}&\textcolor{grey}{28.4}& \textbf{19.8} &\textcolor{grey}{25.6}&\textcolor{grey}{28.9}&\textcolor{grey}{25.9}\\
  \bottomrule
\end{tabular}
\vspace{-2mm}
\caption{Comparison with ViLD on LVIS v1 dataset. * denotes our re-implementation version, see Section~\ref{sec:details} for the details. The frequent and common  classes are used as the base classes, while the rare classes are held out as the novel classes. $\text{AP}_r$ is the main evaluation metric for open-world object detection.}
\vspace{-2mm}
\label{tab:openvocab}
\end{table*}

\begin{table*}[th]
  \small
  \centering
  \begin{tabular}{l|cc|cccccc|cccccc}
  \toprule
 \multirow{2}{*}{Method}&\multicolumn{2}{c|}{Pascal VOC}&\multicolumn{6}{c|}{COCO}&\multicolumn{6}{c}{Objects365}\\
  &AP$_{50}$&AP$_{75}$&AP&AP$_{50}$&AP$_{75}$&AP$_s$&AP$_m$&AP$_l$&AP&AP$_{50}$&AP$_{75}$&AP$_s$&AP$_m$&AP$_l$\\
  \midrule
Supervised&78.5&49.0&46.5&67.6&50.9&27.1&67.6&77.7&25.6&38.6&28.0&16.0&28.1&36.7\\
  \midrule
  ViLD*~\cite{gu2021zero}&73.9&\textbf{57.9}&34.1&52.3&36.5&21.6&38.9&46.1&11.5&17.8&12.3&4.2&11.1&17.8\\
  DetPro &\textbf{74.6}&\textbf{57.9}&\textbf{34.9}&\textbf{53.8}&\textbf{37.4}&\textbf{22.5}&\textbf{39.6}&\textbf{46.3}&\textbf{12.1}&\textbf{18.8}&\textbf{12.9}&\textbf{4.5}&\textbf{11.5}&\textbf{18.6}\\

\bottomrule
\end{tabular}
\vspace{-2mm}
\caption{We evaluate the LVIS-trained model on Pascal VOC test set, COCO validation set and Object365 validation set.}
\vspace{-4mm}
\label{tab:generalization}
\end{table*}

We briefly describe the training process as in~\cite{gu2021zero}: for each region proposal generated by RPN, we pass it through the text head and image head respectively to extract two RoI features for subsequent loss computations. There are two losses: for the text head, cosine similarities between RoI features and base class embedding is computed for classification and a standard cross entropy loss $\mathcal{L}_{\text{text}}$ is adopted. As for the image head branch, we crop and resize proposals generated by the RPN, and feed them into $\mathcal{I}(\cdot)$ to generate image embedding.
A L1 loss (\ie $\mathcal{L}_{\text{image}}$) is applied to minimize the distance between image embedding and the corresponding RoI feature extracted by image head. The generation of image embedding can be performed offline by using a pre-trained RPN.
The overall classification loss is the weighted sum of $\mathcal{L}_{\text{text}}$ and $\mathcal{L}_{\text{image}}$. In addition, we replace the second-stage class-specific bounding box regression and mask prediction layers with class-agnostic modules. A standard regression loss and mask prediction loss are also utilized during training.

\noindent\textbf{Inference ViLD with DetPro.} At inference stage, we use Eq.~\ref{eq:ctx} to generate prompt representations for both base and novel classes, and class embedding is extracted by feeding prompt representations into $\mathcal T({\cdot})$. Thanks to the shared context vectors, prompt representations optimized by DetPro can be well generalized to novel classes though trained on only base classes. Given a test image $x$, RPN first generates a set of proposals. We pass each proposal through the text head and the image head to extract two RoI features (see Figure~\ref{fig:vild}). For each one, we compute its cosine similarities to all class embedding to obtain confidence scores. The final probability for $x$ is the geometric mean of two confidence scores. 

\vspace{-2mm}
\section{Experiment}
\label{sec:exp}
\subsection{Dataset and Evaluation Metrics}
We conduct our main experiments on LVIS v1~\cite{gupta2019lvis} dataset. DetPro and its open-vocabulary object detector are trained on LVIS base classes. We evaluate our approach on LVIS novel classes. Meanwhile, we conduct transfer experiments to demonstrate generalization ability of our approach, and evaluate our LVIS-trained model on Pascal VOC~\cite{everingham2010pascal} test set, COCO~\cite{lin2014microsoft} validation set and Objects365~\cite{shao2019objects365} validation set.

\noindent\textbf{LVIS V1 Dataset.} 
LVIS v1 is a large-scale object detection dataset with a long-tail data distribution. It divides the categories into `frequent', `common', `rare' according to their appearing frequency in the training set. 
Following ViLD~\cite{gu2021zero}, The frequent and common classes are used as the base classes (866 classes), while the rare classes are held out as the novel classes (337 classes).

\noindent\textbf{Pascal VOC Dataset.}
Pascal VOC is a collection of datasets (including VOC2007 and VOC2012) for object detection which contains 20 object categories.

\noindent\textbf{COCO.}
COCO is a standard dataset comprising 80 categories of common objects in natural context. It contains $\sim$118k images with bounding box and instance segmentation annotations. Following ViLD~\cite{gu2021zero}, the instance masks are not computed.

\noindent\textbf{Objects365 Dataset.}
Objects365 is a brand-new large-scale object detection dataset with 365 categories and high-quality bounding box annotations.

\noindent\textbf{Evaluation Metrics.}
We use average precision (AP) to evaluate the performance of object detection and instance segmentation. For LVIS experiments, AP$_r$ is the main indicator, we report the results of AP$_f$ and AP$_c$ as well. While for transfer experiments on Pascal VOC, COCO and Objects365, we use AP, AP$_{50}$, AP$_{75}$, AP$_{s}$, AP$_{m}$ and AP$_{l}$ as the evaluation metrics.

\subsection{Implementation Details}
\label{sec:details}
\noindent\textbf{DetPro.} Unless otherwise specific, we use the following settings of DetPro: context length of 8; class token in the end; 10\% of the background proposals; implicitly model background by Eq.~\ref{eq:softBG}. The context vectors are initialized by drawing from a zero-mean Gaussian distribution of standard deviation 0.02. We choose SGD optimizer with an initial learning rate of 0.002 which is decayed by the cosine annealing rule. We train our DetPro for 6 epochs.

\noindent\textbf{ViLD and Object Detector.} We use the Mask R-CNN with ResNet-50 and FPN as our detector. The model is trained on 8 GPUs with 2 images per GPU. Synchronized batch normalization is used. We use SGD as the optimizer, the momentum and the weight decay are set to 0.9 and 0.00003, respectively. For the comparison with state-of-the-art methods, our detector is trained for 20 epochs, and we train 12 epochs for ablation studies. The learning rate is initialized as 0.02, it is divided by 10 at 16-th epoch and 8-th epoch for the 20-epoch and 12-epoch schedule, respectively. A warm up step with learning rate of 0.001 is performed for the first 500 iterations. We re-implement the ViLD~\cite{gu2021zero}, named ViLD*, by replacing the pre-trained ResNet-50 with self-supervised pre-trained SoCo~\cite{wei2021aligning} to reduce the huge training cost. In the original implementation of ViLD, the whole training process takes up to 180,000 iterations with batchsize of 256, approximately 460 epochs, which is unaffordable. In our re-implementation, the training epoch is reduced from 460 to 20 while comparable performance is achieved.

\noindent\textbf{Vision-Language Model.}
We use the publicly available CLIP\footnote{https://github.com/openai/CLIP} as pre-trained vision-language model. We adopt the ViT-B/32 as the image encoder.

\subsection{Main Results}
\label{sec:main result}
\noindent\textbf{Experiment on LVIS v1 Dataset.} Table~\ref{tab:openvocab} shows the comparison with ViLD on the LVIS v1 dataset. Our re-implementation version of ViLD (denoted as ViLD *) achieves comparable $\text{AP}$ compared with the original implementation, while reducing the training epochs from 460 to 20. Note our performance on $\text{AP}_r$ is even slightly higher while the high $\text{AP}_c$ and $\text{AP}_f$ of original ViLD is owed to the large scale jittering augmentation with long training schedule (approximately 460 epochs). Our DetPro improves the baseline ViLD* by +3.4 $\text{AP}_r$ on object detection, and +3.0 $\text{AP}_r$ on instance segmentation, respectively.

\noindent\textbf{Transfer to Other Datasets.}
Following ViLD~\cite{gu2021zero}, we conduct experiments on transferring LVIS-trained DetPro to other datasets, namely Pascal VOC 2007 test set, COCO validation set and Objects365 v1 validation set, by directly replacing the class tokens. As is shown in Table~\ref{tab:generalization}, Our DetPro improves the baseline ViLD* on all three datasets on Pascal VOC, COCO and Objects365, demonstrating the effectiveness and generalization of our DetPro.

\subsection{Ablation Study}
\label{sec:ablation}
We use LVIS setting, where our model is trained on the LVIS base classes and evaluated on the LVIS rare classes, for all ablation studies. We report the results of instance segmentation. AP$_r$ is used as the main indicator to evaluate the generalization of DetPro.

\begin{table}[t]
  \small
  \centering
  \begin{tabular}{l|cccc}
  \toprule
  Strategy &AP$_r$&AP$_c$&AP$_f$&AP\\
  \midrule
  DetPro w/o BG & 16.9& 25.1 & 27.7 &24.7 \\
  \midrule
  DetPro-LearnableBG&15.3&\textbf{25.4}&27.9&24.6\\
  DetPro-SoftBG&\textbf{19.1}&\textbf{25.4}&\textbf{28.2}&\textbf{25.4}\\
\bottomrule
\end{tabular}
\vspace{-2mm}
\caption{Ablation study on different strategies of involving negative proposals into our DetPro.}
\vspace{-4mm}
\label{tab:variants}
\end{table}

\noindent{\bf Different Ways for Background Interpretation.}
As described in Section~\ref{sec:detection_prompt}, we introduce two strategies to include negative (background) proposals, namely DetPro-SoftBG (Eq.~\ref{eq:softBG}) and DetPro-LeanableBG (Eq.~\ref{eq:neg_prob},\ref{eq:negloss}). Table~\ref{tab:variants} compares two variants with a baseline named DetPro w/o BG, in which neither negative set nor negative loss are used. DetPro-SoftBG ourperforms the baseline by +2.2 AP$_r$, which demonstrates the importance of involving background in a proper way. We observe that DetPro-LearnableBG is worse than the baseline by -1.6 AP$_r$. We conjecture that background content may vary a lot, learning an explicit background embedding to let all negative proposals be close to it, which can not be sufficient.

\noindent\textbf{Number of Negative Proposals.}
We already demonstrate the importance of involving negative proposals into our DetPro, but how many of them should we use in training? Table~\ref{tab:bgsamples} shows the study. The result on AP$_r$ consistently declines with the increasing of negative samples.
Since negative samples are significantly more than positive ones, reducing negatives can avoid bias towards background as well as speed up training. Our default is 10.

\begin{table}[t]
  \small
  \centering
  \begin{tabular}{c|cccc}
  \toprule
  Background proposals&AP$_r$&AP$_c$&AP$_f$&AP\\
  \midrule
  \textbf{10\%}&\textbf{19.1}&25.4&28.2&\textbf{25.4}\\ 
  30\%&18.3&\textbf{25.6}&\textbf{28.4}&\textbf{25.4}\\
  50\%&17.8&\textbf{25.6}&\textbf{28.4}&\textbf{25.4}\\
  100\%&17.6&25.1&28.2&25.0\\
\bottomrule
\end{tabular}
\vspace{-2mm}
\caption{Ablation on number of background proposals involved in DetPro training.}
\vspace{-3mm}
\label{tab:bgsamples}
\end{table}

\begin{table}[t]
  \small
  \centering
  \begin{tabular}{ccc|cccc}
  \toprule
  GT & FG & BG &AP$_r$&AP$_c$&AP$_f$&AP\\
  \midrule
  \checkmark & &  &15.3&\textbf{25.4}&27.9&24.6\\
  \checkmark & \checkmark &  &16.9&25.1&27.7&24.7\\
  \checkmark & & \checkmark &17.7&25.3&\textbf{28.2}&25.1\\
  \checkmark& \checkmark&\checkmark &\textbf{19.1}&\textbf{25.4}&\textbf{28.2}&\textbf{25.4}\\
\bottomrule
\end{tabular}
\vspace{-2mm}
\caption{Ablation study on the involvement of different training data. 'GT': ground-truth; `FG': foreground; `BG': background.}
\label{tab:fg_for_train}
\vspace{-4mm}
\end{table}

\noindent{\bf Involvement of Different Training Data.}
We study various combinations of training data in Table~\ref{tab:fg_for_train}. Our default setting, \ie, including ground-truth, foreground proposals and background proposals, yields the best performance among others. Eliminating either foreground proposals or background proposals from training data leads to performance degradation. Using only ground-truth for training degenerates to CoOp~\cite{zhou2021learning}.

\noindent{\bf Context Grading and Prompt Representation Ensemble.}
Here we study the effects of prompt representation ensemble as shown in Table~\ref{tab:positive_proposals}. As described in Section~\ref{sec:detection_prompt}, we divide the positive proposals of the IoU range $[a,b]$ into $K$ disjoint groups with an IoU interval of $t$. Then class embedding from $K$ learned DetPro are ensembled. From the table we observe that our DetPro with ensemble strategy consistently improve the performance over their non-ensemble counterparts, \eg `Ensemble (0.5:1.0:0.1)' outperforms `IoU range = [0.5-1.0]' by +3.0 AP$_r$. The main improvements come from the novel classes.

\begin{table}[t]
  \small
  \centering
  \begin{tabular}{c|cccc}
  \toprule
  IoU range &AP$_r$&AP$_c$&AP$_f$&AP\\
  \midrule
  0.5-0.6 & 17.3&25.3&28.2&25.0\\
  0.6-0.7 &18.0&25.4&28.1&25.4\\
  0.7-0.8 &17.2&25.4&28.3&25.1\\
  0.8-0.9 &17.3&24.9&28.2&24.9\\
  0.9-1.0 &17.2&25.2&28.3&25.0\\
  \midrule
0.5-1.0 &16.1&25.7&28.3&25.1\\
0.6-1.0 &17.2&25.4&28.9&25.3\\
0.7-1.0 &16.8&25.0&28.3&25.1\\
0.8-1.0 &17.2&25.2&28.4&25.1\\
  \midrule
  \midrule
\textbf{Ensemble (0.5:1.0:0.1)} &\textbf{19.1}&25.4&28.2&25.4\\
Ensemble (0.6:1.0:0.1) &18.4&25.2&28.2&25.2\\
Ensemble (0.7:1.0:0.1) &18.7&\textbf{25.8}&\textbf{28.3}&\textbf{25.5}\\
Ensemble (0.8:1.0:0.1) &18.2&25.3&28.1&25.2\\
  
\bottomrule
\end{tabular}
\vspace{-2mm}
\caption{The effects of prompt representation ensemble. `Ensemble (0.5:1.0:0.1)' represents we divide the positive proposals of the IoU range [0.5-1.0] into 5 disjoint groups with an IoU interval of 0.1. Then we use each group to train a separate DetPro and perform ensemble on 5 learned models.} 
\vspace{-2mm}
\label{tab:positive_proposals}
\end{table}

\noindent{\bf Context Length.} We study the effects of using different context lengths $L$. We vary the length from 4 to 8 to 16 and Table~\ref{tab:context length} shows the study. CoOp~\cite{zhou2021learning} has shown that using longer prompt can lead to better performance on close-vocabulary image classification task. We obtain the same conclusion from the performance of base classes (AP$_c$ and AP$_f$). However, it does not hold true for novel classes, suggesting that longer prompt may cause over-fitting to base categories. We set context length as 8 by default.

\begin{table}[t!]
  \small
  \centering
  \begin{tabular}{c|cccc}
  \toprule
  Length&AP$_r$&AP$_c$&AP$_f$&AP\\
  \midrule
  4&18.7&24.9&28.2&25.1\\
  \textbf{8}&\textbf{19.1}&\textbf{25.6}&\textbf{28.3}&25.2\\
  16&17.7&\textbf{25.6}&\textbf{28.3}&\textbf{25.3}\\
\bottomrule
\end{tabular}
\vspace{-2mm}
\caption{Ablation study on context lengths.}
\vspace{-3mm}
\label{tab:context length}
\end{table}

\noindent{\bf Position of Class Token.}
Table~\ref{tab:position} studies inserting class token into different positions, namely front, middle and end, of the prompt representations. Generally, the best position depends on the dataset~\cite{zhou2021learning}. In our experiment, positioning class token in the end achieves the best performance.
\begin{table}[t]
  \small
  \centering
  \begin{tabular}{l|cccc}
  \toprule
  Position&AP$_r$&AP$_c$&AP$_f$&AP\\
  \midrule
  Front&16.4&24.5&\textbf{28.3}&24.6\\
  Middle&18.0&25.1&\textbf{28.3}&25.1\\
  \textbf{End}&\textbf{19.1}&\textbf{25.4}&28.2&\textbf{25.4}\\
\bottomrule
\end{tabular}
\vspace{-2mm}
\caption{Ablation study of inserting class token into different positions of prompt representation.}
\label{tab:position}
\vspace{-5mm}
\end{table}

\vspace{-1mm}
\subsection{Visualization}
\begin{figure}[t]
    \centering
    \begin{subfigure}{0.45\linewidth}
        \includegraphics[width=\textwidth]{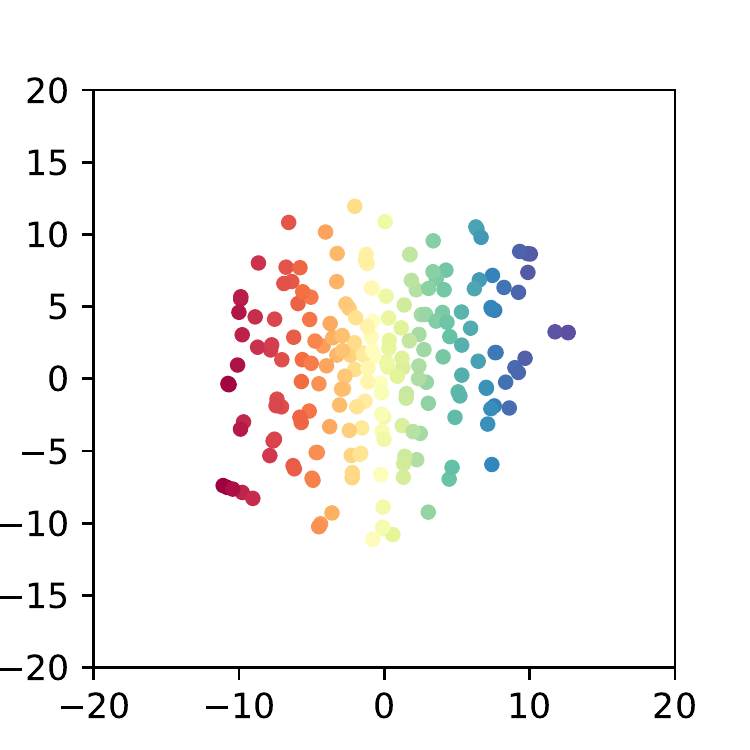}
        \caption{Prompt engineering (base).}
    \end{subfigure}
    \quad
    \begin{subfigure}{0.45\linewidth}
        \includegraphics[width=\textwidth]{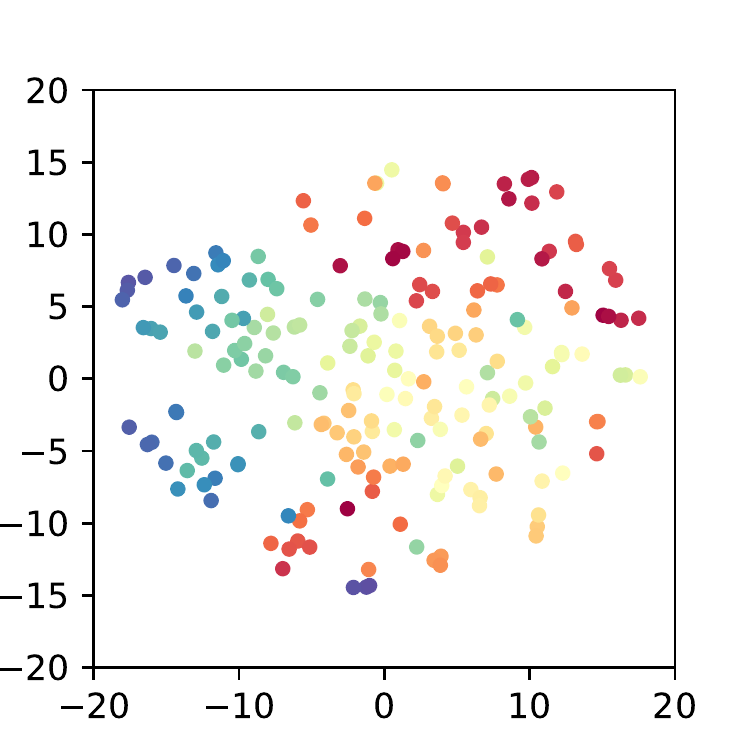}
        \caption{DetPro  (base).}
    \end{subfigure}
        \begin{subfigure}{0.45\linewidth}
        \includegraphics[width=\textwidth]{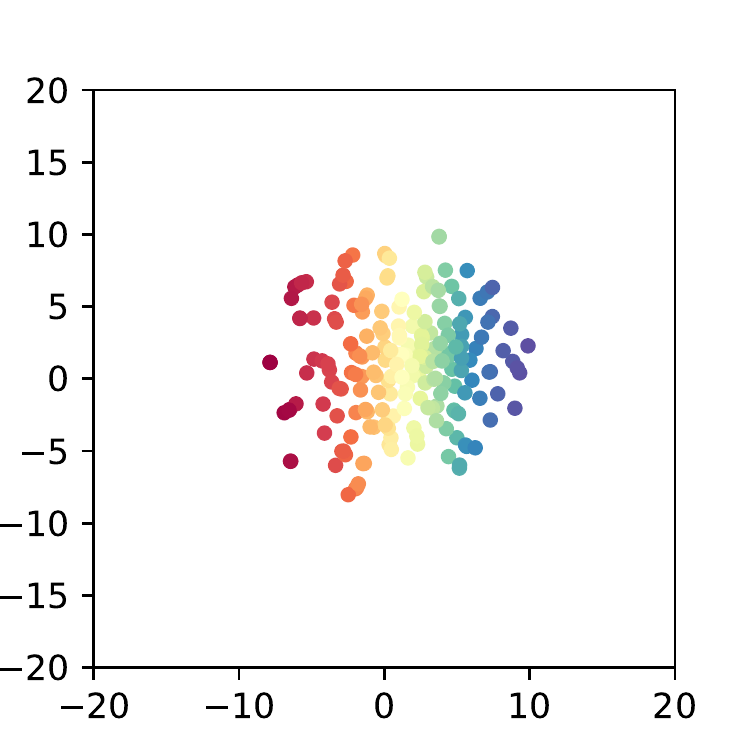}
        \caption{Prompt engineering  (novel).}
    \end{subfigure}
    \quad
    \begin{subfigure}{0.45\linewidth}
        \includegraphics[width=\textwidth]{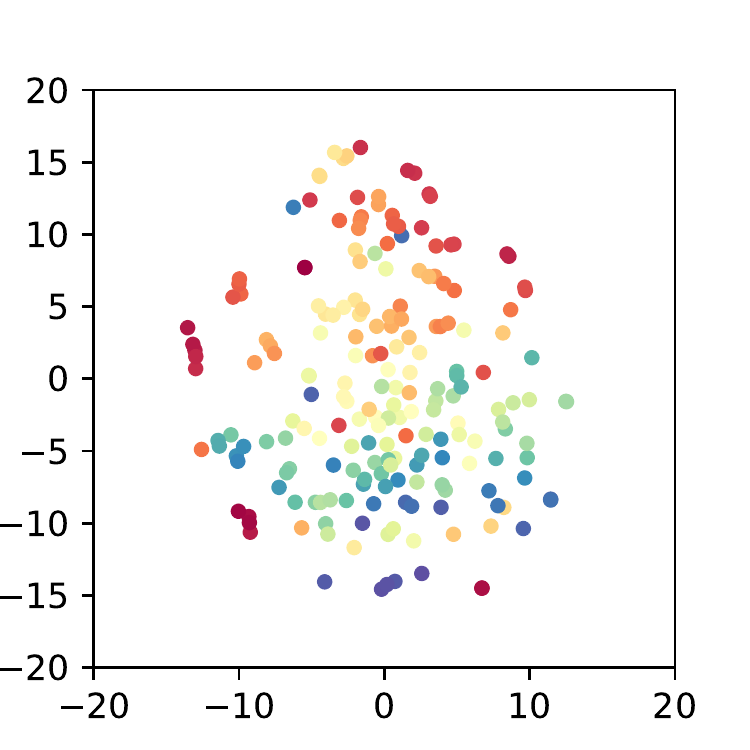}
        \caption{DetPro (novel).}
    \end{subfigure}
    \vspace{-1mm}
    \caption{We randomly select 200 base classes and 200 novel classes from LVIS dataset and use t-SNE to visualize the class embedding generated by our DetPro and the classical prompt engineering. (a) base class embedding generated by prompt engineering; (b) base class embedding generated by DetPro; (c) novel class embedding generated by prompt engineering; (d) novel class embedding generated by DetPro. Each point denotes a category. Class embedding generated by our method is more discriminative in the embedding space, which attributes to the involvement of background proposals during training.}
    \label{fig:t-sne}
    \vspace{-4mm}
\end{figure}

\vspace{-1mm}
To further demonstrate the importance of involving both foreground and background proposals in detection-oriented prompt representation learning. We randomly  select 200 base classes and 200 novel classes from LVIS dataset and use t-SNE to visualize the class embedding generated by DetPro and prompt engineering as shown in Figure~\ref{fig:t-sne}. We observe that the class embedding generated by DetPro is more discriminative in the embedding space, this superior property indicates they are more capable of being region classifiers for open-vocabulary object detector.

\section{Conclusion}
\vspace{-1mm}
\label{sec:conclusion}
In this paper, we propose a novel method named detection prompt (DetPro), aiming to learn continuous prompt representations for open-vocabulary object detection based on the pre-trained vision-language model. Different from the previous classification-oriented prompt learning method, DetPro presents a background interpretation scheme to include negative proposals in images into the training, and a context grading scheme to separate positive proposals in image foreground for tailored prompt training. We assemble DetPro with ViLD and conduct a series of studies to demonstrate the importance of involving both proposals in both foreground and background in prompt representation learning for open-vocabulary object detection. Experiments on LVIS and transfer learning on Pascal VOC, COCO, Objects365 demonstrate the effectiveness and generalization ability of our approach.

\subsection*{Acknowledgements}
This work was partially supported  by  National Nature Science Foundation of China (No. 61836004, 61828602), National Key Research and Development Program of China (Grant No. 2021ZD0200300), National Key R\&D Program of China (2018AAA0102600), and Beijing Academy of Artificial Intelligence (BAAI).

\appendix

\section{More Experiments and Analysis}
\noindent\textbf{A Variant of DetPro.}
Following ViLD~\cite{gu2021zero}, we present a variant of DetPro named DetPro-text, and compare it with ViLD-text. In the DetPro-text, we remove the image head and only use a text head for training and inference. We use the LVIS setting as described in Section~5.3. Table~\ref{tab:variant_detpro} shows the comparison.

\begin{table}[t]
  \small
  \centering
  \begin{tabular}{c|cccc}
  \toprule
 Method&AP$_r$&\textcolor{grey}{AP$_c$}&\textcolor{grey}{AP$_f$}&\textcolor{grey}{AP}\\
  \midrule
ViLD-text*\cite{gu2021zero}&12.1&\textcolor{grey}{24.2}&\textcolor{grey}{28.9}&\textcolor{grey}{23.9}\\
DetPro-text&\textbf{14.2}&\textcolor{grey}{23.9}&\textcolor{grey}{28.9}&\textcolor{grey}{24.2}\\

\bottomrule
\end{tabular}
\caption{We compare our DetPro-text with the ViLD-text. * denotes our re-implementation version, see Section~5.2 for the details.}
\label{tab:variant_detpro}
\end{table}

\noindent\textbf{Using Different Number of Positive Samples for Training.}
We also study the effects of using different number of positive samples in DetPro training as shown in Table~\ref{tab:possamples}. The LVIS setting is adopted in this study. We observe that using all positive samples results in the best generalization performance on novel classes.
\begin{table}[t]
  \small
  \centering
  \begin{tabular}{c|cccc}
  \toprule
  Positive samples (\%)&AP$_r$&AP$_c$&AP$_f$&AP\\
  \midrule
  10&18.2&25.4&28.2&25.3\\ 
  30&18.4&25.1&28.2&25.1\\
  50&18.8&25.4&28.2&25.4\\
  \textbf{100}&\textbf{19.1}&\textbf{25.4}&\textbf{28.2}&\textbf{25.4}\\
\bottomrule
\end{tabular}
\caption{Ablation study of using different number of positive samples for DetPro training.}
\label{tab:possamples}
\end{table}

\noindent\textbf{Accuracy of Proposal Classification.} 
In our DetPro, we first optimize the prompt representations then feed them into CLIP text encoder to generate class embedding as classifiers of the detector. Here we report the image proposal classification accuracy on the LVIS dataset
to demonstrate the effectiveness of our approach. Concretely, given a set of proposals generated by RPN, we resize each
proposal to the size of $224 \times 224$ and feed it into the CLIP image encoder to extract its image embedding, then we compute the similarities between the image embedding and all class embedding to predict its class. We compare our approach with the prompt engineering. Table~\ref{tab:cropped t1 acc} and Table~\ref{tab:cropped t5 acc} show top-1 and top-5 accuracy, respectively. Remarkable improvements are observed on both base classes and novel classes, indicating that the prompt representations learned by our DetPro are also beneficial to the open-vocabulary image classification task. 

\begin{table}[t]
  \centering
  \begin{tabular}{l|c|c}
  \bottomrule
  Method&Base class& Novel class\\
  \hline
  Prompt engineering&20.1&17.7\\
  DetPro& \textbf{24.4}& \textbf{21.7}\\
\bottomrule
\end{tabular}
\caption{Top-1 accuracy of proposal classification.}
\label{tab:cropped t1 acc}
\end{table}

\begin{table}[t]
  \centering
  \begin{tabular}{l|c|c}
  \bottomrule
  Method&Base class& Novel class\\
  \hline
  Prompt engineering&39.3&37.2\\
  DetPro&\textbf{49.0} & \textbf{40.3}\\
\bottomrule
\end{tabular}
\caption{Top-5 accuracy of proposal classification.}
\label{tab:cropped t5 acc}
\end{table}

\noindent\textbf{Assembling the Well-trained ViLD with DetPro.}
In Section~4.3 of the main paper, we use class embedding generated by DetPro for the ViLD training and inference. In this study, we first use class embeddings generated by prompt engineering as classifiers of the detector to train ViLD, and assemble the well-trained ViLD with our DetPro for inference, by simply replacing the original class embedding in the image head with the ones generated by our DetPro. It can be seen in Table~\ref{tab:replace} that simply assembling the original ViLD with DetPro trained with different ensemble strategies (see Table 6 of the main paper) already shows non-negligible improvements on novel classes.

\begin{table}[h]
  \small
  \centering
  \begin{tabular}{c|cccc}
  \toprule
  Method&AP$_r$&AP$_c$&AP$_f$&AP\\
  \midrule
  ViLD*&16.8&25.6&28.5&25.2\\
  \midrule
  DetPro-Ensemble(0.5:1.0:0.1)&18.1&25.7&28.3&25.4\\
  DetPro-Ensemble(0.6:1.0:0.1)&18.0&25.4&28.2&25.2\\
  DetPro-Ensemble(0.7:1.0:0.1)&18.0&25.4&28.2&25.3\\
  DetPro-Ensemble(0.8:1.0:0.1)&17.9&25.7&28.3&25.4\\
\bottomrule
\end{tabular}
\caption{Assembling the well-trained ViLD with DetPro trained under different settings outperforms the original ViLD.}
\label{tab:replace}
\end{table}

\begin{figure}[t]
    \centering
    \begin{subfigure}{0.45\linewidth}
        \includegraphics[width=\textwidth]{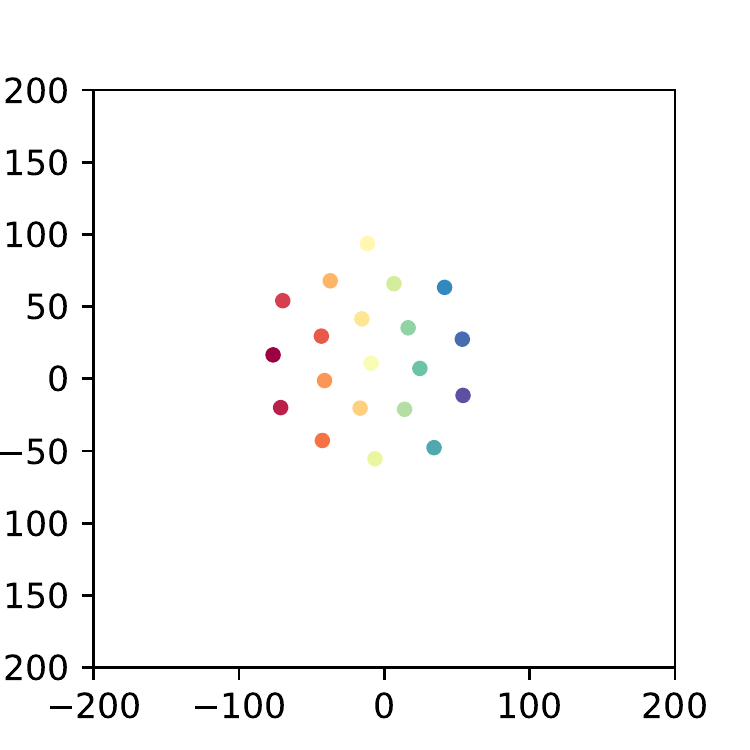}
        \caption{Prompt engineering.}
    \end{subfigure}
    \quad
    \begin{subfigure}{0.45\linewidth}
        \includegraphics[width=\textwidth]{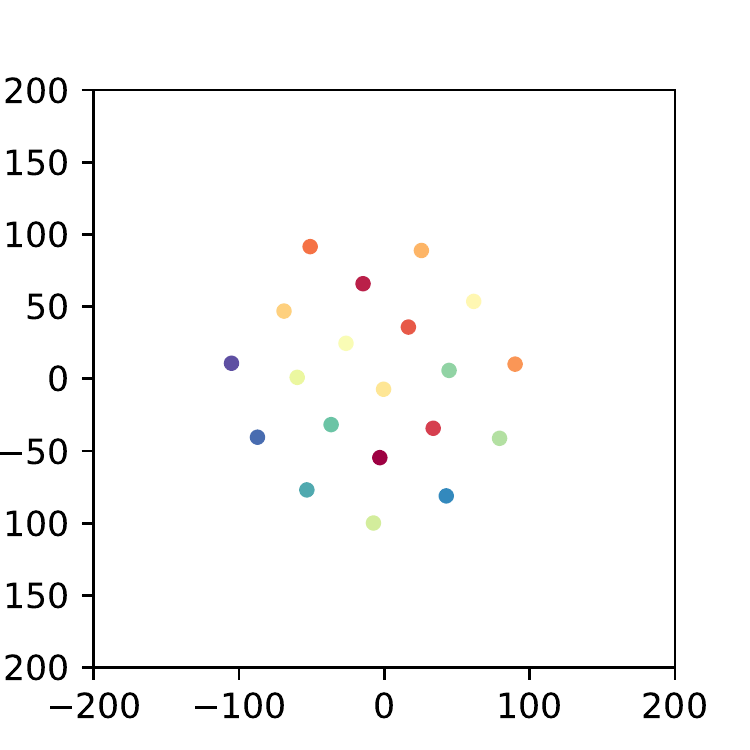}
        \caption{DetPro.}
    \end{subfigure}
    \caption{T-SNE visualization for Pascal VOC dataset.}
    \label{fig:voc}
    \vspace{-2mm}
\end{figure}

\begin{figure}[t]
    \centering
    \begin{subfigure}{0.45\linewidth}
        \includegraphics[width=\textwidth]{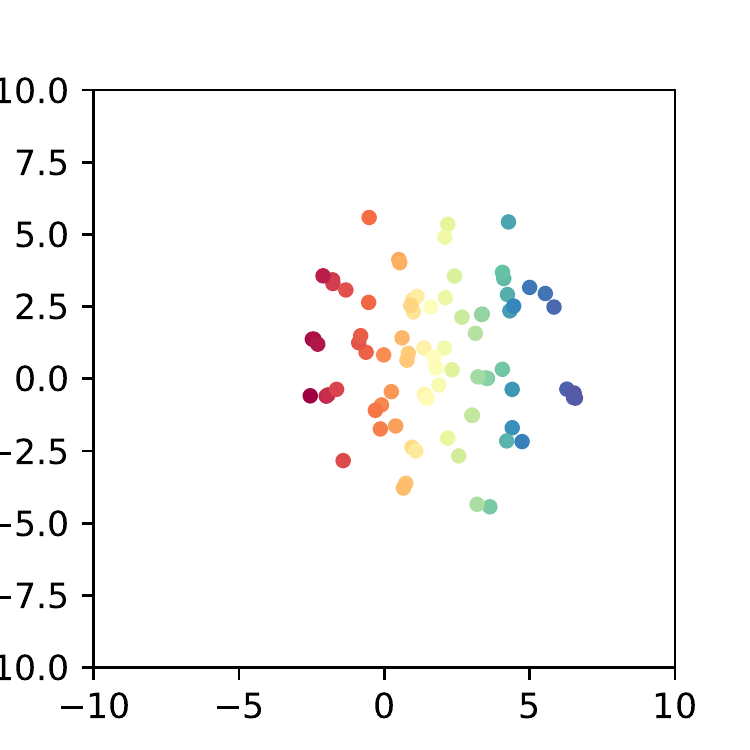}
        \caption{Prompt engineering.}
    \end{subfigure}
    \quad
    \begin{subfigure}{0.45\linewidth}
        \includegraphics[width=\textwidth]{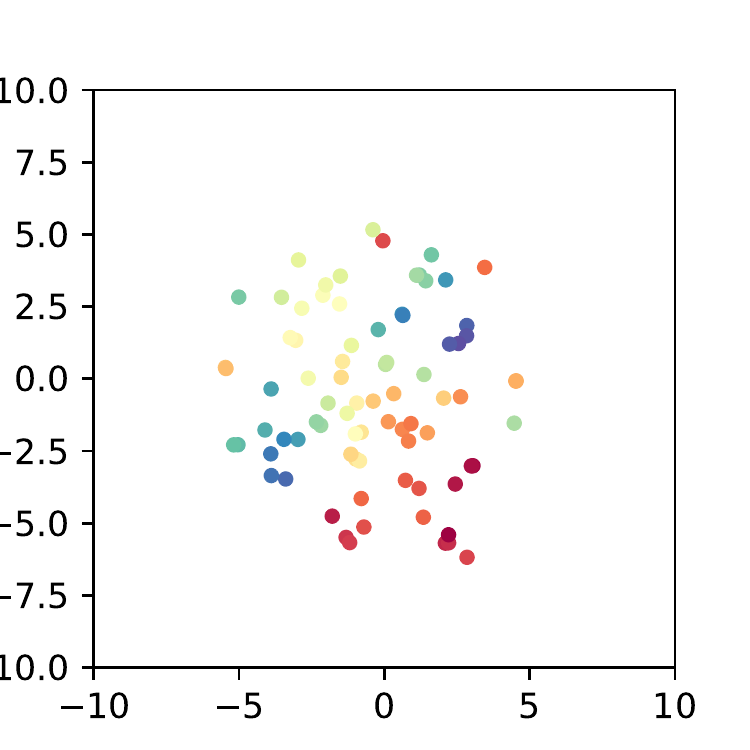}
        \caption{DetPro.}
    \end{subfigure}
    \caption{T-SNE visualization for COCO dataset.}
    \label{fig:coco}
    \vspace{-2mm}
\end{figure}

\begin{figure}[t]
    \centering
    \begin{subfigure}{0.45\linewidth}
        \includegraphics[width=\textwidth]{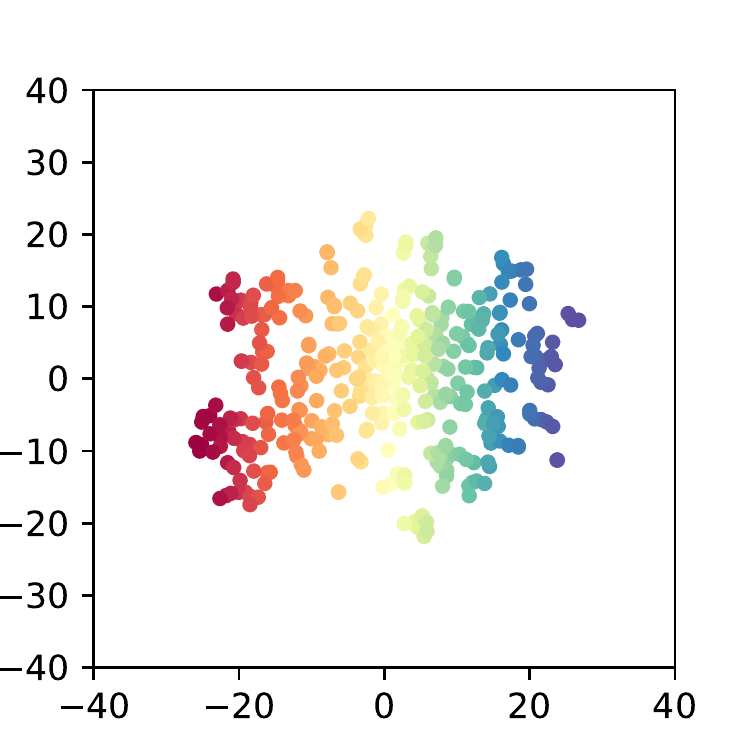}
        \caption{Prompt engineering.}
    \end{subfigure}
    \quad
    \begin{subfigure}{0.45\linewidth}
        \includegraphics[width=\textwidth]{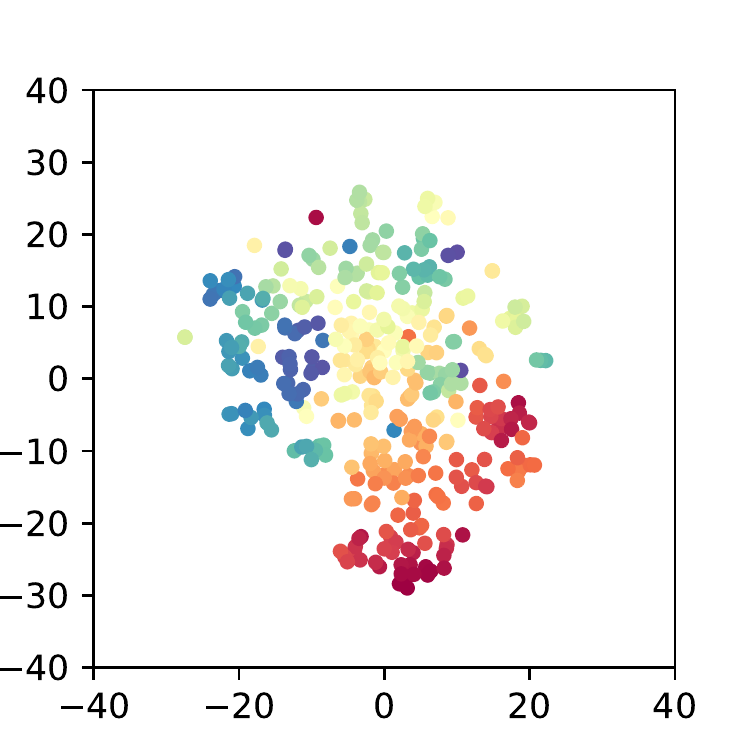}
        \caption{DetPro.}
    \end{subfigure}
    \caption{T-SNE visualization for Objects365 dataset.}
    \label{fig:objects365}
    \vspace{-2mm}
\end{figure}

\noindent\textbf{T-SNE Visualization for Transferred Datasets.} In Section 5.5, we generate the class embedding for the LVIS dataset and show the t-SNE figure. Here we use t-SNE to visualize the class embedding generated by our DerPro and prompt engineering on transferred datasets including Pascal VOC, COCO, and Objects365. Figure~\ref{fig:voc}-\ref{fig:objects365} show the comparison. We observe the same phenomenon that the class embedding generated by DetPro is more discriminative in the embedding space, which further validates their suitability serving as the region classifiers for open-vocabulary object detection.

\section{More Implementation Details}
\noindent\textbf{More Details of Our Open-world Object Detector.}
We use multi-scale training with the size of (1333, 640), (1333, 672), (1333, 704), (1333, 736), (1333, 768), (1333, 800). For RPN, we apply an NMS with a threshold of 0.7 and generate a maximum of 1000 proposals. We apply a class-agnostic NMS with a threshold of 0.5 on the final predictions and set the maximum number of output bounding boxes to 300.

\noindent\textbf{More Details of DetPro Training.}
We set the batch size as 512. We use a cross-entropy loss with a temperature parameter of 0.01.
{\small
\bibliographystyle{ieee_fullname}
\bibliography{egbib}
}

\end{document}